\documentclass[11pt]{article}

  \usepackage[preprint]{acl}

 \usepackage{times}
\usepackage{latexsym}
\usepackage{amsmath}
 \usepackage[T1]{fontenc}
   \usepackage{booktabs}
 \usepackage[utf8]{inputenc}
\usepackage{multirow}
   \usepackage{microtype}
\usepackage{tikz}
   \usepackage{inconsolata}
\usepackage{graphicx}
\usepackage{fancyvrb}
\usepackage{fvextra}

\title{Key Point Analysis Needs Structure Recovery: Task Definition, Dataset Diagnosis, and a Structure-Aware Benchmark}

\author{Zhiqiang Shi \\
  Department of Informatics \\
  King's College London \\
  \texttt{zhiqiang.shi@kcl.ac.uk} \\\And
  Oana Cocarascu \\
  Department of Informatics \\
  King's College London \\
  \texttt{oana.cocarascu@kcl.ac.uk} \\}

\begin{document}
\maketitle
\begin{abstract}
Key Point Analysis (KPA) aims to identify a concise set of key points that summarize a collection of arguments together with their prevalence. We argue that KPA is fundamentally a structured prediction problem that requires recovering semantic groupings, generating representative key points, ensuring coverage, and estimating prevalence. Under this formulation, we show that existing KPA benchmarks suffer from limitations in grouping quality, redundancy, coverage, and argument–key point mappings, causing ceiling violation and selection failure in reference-based evaluation. To support future research on true KPA, we introduce a structure-aware, distribution-sensitive benchmark built via a human-in-the-loop re-annotation. Human and LLM evaluations consistently show that the resulting structures yield more coherent groupings, higher-quality key points, better coverage, and more reliable prevalence estimates than existing annotations. We further release several annotation resources to support research on KPA evaluation, argument–key point matching, explainable KPA, and LLM-as-a-judge methodologies, and  outline a research agenda for true KPA. \footnote{The benchmark can be found at: \url{https://github.com/zqshizhiqiang/kpa_dataset}.}

\end{abstract}

\section{Introduction}
     Key Point Analysis (KPA) aims to identify a concise set of key points (KPs) that capture the main arguments in a collection of opinionated texts together with their prevalence  \cite{bar-haim-etal-2020-arguments, friedman-etal-2021-overview}. By providing both a qualitative summary of main viewpoints and a quantitative view of their distribution, KPA enables applications in opinion mining, policy analysis, and large-scale survey interpretation \cite{friedman-etal-2021-overview}.

The original definition of KPA extends beyond summarization~\cite{bar-haim-etal-2020-arguments, friedman-etal-2021-overview}:  it requires identifying prominent and concise KPs that reflect the underlying argument corpus while accurately modeling their prevalence.  We argue that this objective is fundamentally a structured prediction problem.
Specifically, KPA involves: \textbf{semantic grouping} (defining the \textit{group structure} by organizing arguments into semantically coherent clusters), \textbf{KP generation} (providing a \textit{semantic representation} of each group by generating an abstract KP), \textbf{coverage} (ensuring a \textit{complete and non-redundant} representation of the structure), and \textbf{prevalence} (defining a \textit{distribution over the structure} by estimating the relative frequency of each group within the corpus).
 Under this formulation, KPA can be viewed as the task of recovering a structured representation of the argument corpus.

Existing KPA datasets such as ArgKP~\cite{bar-haim-etal-2020-arguments}, ArgKP21~\cite{friedman-etal-2021-overview}, and ArgCMV~\cite{gurjar-etal-2025-argcmv} provide the core task components: arguments, KPs, and argument-KP mappings, which
 in principle  define
argument semantic groupings,
KP-based summaries of these groups, and prevalence via supporting-argument counts.     We argue that the main limitation of KPA datasets lies not in their representation,  but in their  annotation process:
while existing datasets use similar representation formats, their annotation strategies limit  evaluation of true KPA, particularly in the alignment between KPs and argument distributions,  coverage, and prevalence estimation.
 To this end, we conduct a structure-aware human evaluation assessing annotated KPs and alternative KP sets independently on the requirements of true KPA (i.e. semantic grouping, KP quality, and coverage)
 and
show that alternative KP sets
consistently outperform  the ground-truth annotations.
 This reveals  a \textbf{ceiling violation}, where the reference annotations are not optimal,  and a \textbf{selection failure}, where evaluation based on similarity to annotations favors the annotated KPs, as model outputs are scored against them.  Thus, the annotations  achieve the highest possible score by definition, even when alternative
KP sets  better capture the underlying arguments and their semantic organization.

To support research on true KPA, we construct a structure-aware, distribution-sensitive benchmark based on ArgKP21~\cite{friedman-etal-2021-overview}.
To better capture the role of argument distributions in KPA, we create multiple argument subsets for each topic, treating each subset as an independent KPA instance, and re-annotate them using a human-in-the-loop process. The new benchmark  reflects  the requirements of true KPA: semantic grouping, KP generation, coverage, and prevalence estimation.

Our contributions are: \textbf{1)}~We formalize KPA as a structured prediction problem that requires recovering the underlying organization of an argument corpus, including semantic grouping, KP generation, coverage, and prevalence estimation;
 \textbf{2)}~We show that existing KPA benchmarks are misaligned with this formulation and thus  cannot fully evaluate true KPA capabilities;
\textbf{3)}~Through a structure-aware human evaluation, we show that ArgKP21 suffers from a \textbf{ceiling violation} and a \textbf{selection failure}, where reference annotations are not optimal yet favored by similarity-based evaluation over better alternatives;  \textbf{4)}~We release ArgKP-X, a re-annotated, structure-aware benchmark  that enables distribution-sensitive evaluation  and better aligns with  true KPA;
\textbf{5)}~We release additional human and LLM-annotated resources  from the annotation process,  supporting future research on argument-KP matching, coverage analysis, KPA evaluation, explainability, and LLM as-a-judge methodologies.

\section{Related Work}

KPA was introduced by \citet{bar-haim-etal-2020-arguments} and formalized by \citet{friedman-etal-2021-overview} as the task of identifying a concise set of KPs that summarize a collection of arguments together with their relative prevalence. There are only three KPA-specific datasets: ArgKP \cite{bar-haim-etal-2020-arguments},
 ArgKP21 \cite{friedman-etal-2021-overview},
 and ArgCMV \cite{gurjar-etal-2025-argcmv}.  Although these datasets include the components required for KPA (arguments, KPs, and argument-KP mappings for the given topic and stance),
we argue that  they exhibit significant limitations that prevent them from evaluating true KPA.

The ArgMining 2021 shared task \cite{friedman-etal-2021-overview} divided KPA into two sub-tasks: KP generation and KP matching. As a result, most subsequent works \cite{alshomary-etal-2021-key, kapadnis-etal-2021-team, reimer-etal-2021-modern, phan-etal-2021-matching, li-etal-2023-hear} treat KPA as two separate tasks.
 Evaluation methods compare generated KPs against a set of reference KPs.
Existing approaches include  ROUGE-based metrics \citep{li-etal-2023-hear,van-der-meer-etal-2024-empirical},
 semantic similarity measures such as soft-precision, soft-recall, and soft-F1 \citep{li-etal-2023-hear} as well as coverage metrics based on semantic matching \citep{khosravani-etal-2024-enhancing}. More recently, LLMs have been used as judges to assess coverage and redundancy directly \citep{altemeyer2025argument}. However, all existing evaluation approaches  assume that the reference annotations represent the optimal KP set and that outputs more similar to the references are of higher quality. In our work, we show that this assumption does not hold for existing KPA datasets.
\section{Towards True Key Point Analysis } \label{section: true kpa}

\subsection{What is True KPA?}

KPA is the task of producing \textit{a succinct list of the most prominent KPs in the input corpus, along with their relative prevalence}~\cite{friedman-etal-2021-overview}. This definition emphasizes two core objectives:  summarizing arguments into concise KPs  and quantifying their importance through prevalence.
  Since KPs are intended to capture the main ideas expressed by collections of arguments, they must be grounded in these underlying groups rather than defined independently. Moreover, producing a set of prominent KPs with corresponding prevalence requires organizing the argument corpus into semantically coherent groups and summarizing each group in a principled way.

 This implies
that KPA inherently involves recovering a structured representation of the argument space.
 Specifically, KPA involves: \textbf{semantic grouping} which organizes arguments into coherent groups expressing similar underlying ideas, \textbf{KP generation} which provides an abstract representation for each group, \textbf{coverage} which ensures a complete and non-redundant representation of the corpus, and \textbf{prevalence} which quantifies the relative frequency  of each group.

Therefore, evaluating KPA requires datasets that support assessment of the underlying argument structure, rather than only surface-level outputs. In particular, a suitable dataset must provide: reliable semantic groupings that reflect the natural organization of arguments, KPs that are grounded in and representative of these groups, comprehensive yet non-redundant coverage of the corpus, and sufficient information to estimate prevalence from the frequency of the underlying semantic groups.
\subsection{Dataset Limitations}

Given the requirements above, we examine whether existing datasets can evaluate true KPA,
 We consider two types of datasets: designed specifically for KPA (ArgKP \cite{bar-haim-etal-2020-arguments}, ArgKP21 \cite{friedman-etal-2021-overview}, ArgCMV \cite{gurjar-etal-2025-argcmv}) and repurposed for KPA (e.g. Perspectrum \cite{chen-etal-2019-seeing}). While these datasets are structurally compatible, we show that they
fail to capture the requirements of true KPA:  semantic grouping, KP generation, coverage, and prevalence.

We identify three recurring limitations: weak alignment between semantic grouping and KP generation, incomplete coverage of the argument space, and unreliable prevalence estimation.\footnote{Appendix \ref{appendix: dataset limitations} provides a detailed analysis of these datasets.}
In ArgKP and ArgKP21, KPs are created independently of the underlying argument groups, while mappings are established through pairwise matching rather than joint structure recovery. Consequently, arguments assigned to the same KP do not necessarily form coherent semantic groups. Conversely, ArgCMV generates KPs from the full corpus  but does not explicitly recover the underlying group structure. Thus, neither dataset jointly captures  KP generation and semantic grouping.    The datasets also exhibit incomplete coverage of the argument space, e.g.  approximately 9--10\% of arguments in ArgKP21 are not assigned to any KP. Furthermore, prevalence estimates are derived from argument-KP mappings. When semantic groups are incomplete or poorly defined, these estimates cannot faithfully reflect the prominence of ideas in the corpus.

Similar issues arise in datasets repurposed for KPA evaluation. For example, Perspectrum organizes arguments as evidence for predefined claims rather than grouping arguments into shared semantic themes. While structurally compatible with KPA, it does not evaluate whether systems recover the latent argument structure required for true KPA.

Overall, existing benchmarks capture aspects of KPA but do not reliably recover the semantic structure underlying an argument corpus. Consequently, they cannot fully evaluate
 true KPA.

\section{Evaluating the Quality of ArgKP21 Annotations} \label{section: argkp evaluation}

In this section, we investigate whether the ground-truth annotations in ArgKP21 satisfy the structural requirements of true KPA. Rather than treating the annotations as the gold standard, we evaluate their quality with respect to semantic grouping, KP representation, redundancy, and coverage. To this end, we conduct a human evaluation that assesses both ground-truth and LLM-generated KPs under a unified framework. Specifically, we perform three complementary analyses: (i) semantic grouping and KP quality, (ii) redundancy among KPs, and (iii) a targeted examination of arguments labeled as unmatched in the ground-truth annotations. This evaluation design allows us to identify both representational and structural limitations of the dataset without relying on reference-based similarity.

All evaluation tasks are performed by a shared pool of annotators who are required to pass a qualification test to ensure a consistent understanding of semantic equivalence and argument grouping. Ground-truth and LLM-generated KPs are evaluated independently under identical instructions. The LLM outputs are generated using Qwen3-235B  without dataset-specific fine-tuning or extensive prompt optimization.\footnote{ See Appendix \ref{appendix: llm prompts} for the prompt.}

\subsection{Evaluation of Semantic Grouping and Key Point Quality}

\noindent\textbf{Setup.}
The evaluation is conducted on the ArgKP21 test set, which consists of three topics.
Each topic includes \textit{pro} and \textit{con} KP sets along with the associated arguments, as well as ground-truth KP–argument matches.   For each annotation instance, annotators are shown the topic, the stance,  a KP, and all arguments for that topic/stance/KP. Annotators are asked to identify exactly  the arguments that express the KP's underlying reasoning,
 provide a brief justification of the shared reasoning among the selected arguments, and rate the KP's quality as a summary of the selected arguments on a four-point scale (\textit{Poor}, \textit{Fair}, \textit{Good}, \textit{Excellent}).
  We recruit three annotators, each assigned to one topic and evaluating  all KPs for that  topic.
 Model outputs are evaluated identically  by associating each generated KP with a set of arguments from the same topic.  This design allows evaluation  at both \textit{topic level} and \textit{overall level} across all topics.

\noindent\textbf{Metrics.} Based on the annotated results, we evaluate the quality of semantic grouping and KP representation using the following metrics.

\noindent\textit{Global Cluster Precision.}
For each KP $k$, we define cluster precision as
\(
\text{Precision}(k) = \frac{|\mathcal{A}_k|}{|\mathcal{M}_k|}\)
where $\mathcal{M}_k$ denotes the set of arguments originally matched to  $k$ and $\mathcal{A}_k$ denotes the subset selected by annotators as expressing the same reasoning.
 We then aggregate precision across all KPs:
\(
\text{Precision}_{global} =
\frac{\sum_k |\mathcal{A}_k|}{\sum_k |\mathcal{M}_k|}
\).

\noindent\textit{Abstraction Score.}
Annotators rate each KP on a four-point scale (\textit{Poor}, \textit{Fair}, \textit{Good}, \textit{Excellent}), which we map to numerical scores $\{1, 2, 3, 4\}$. We report the average abstraction score across KPs.

\begin{table}[t]
\centering
\small
\begin{tabular}{lcccc}
\toprule
\textbf{Topic}
& \multicolumn{2}{c}{\textbf{Global Precision}}
& \multicolumn{2}{c}{\textbf{Abstraction Score}} \\
\cmidrule(lr){2-3} \cmidrule(lr){4-5}
 & \textbf{GT} & \textbf{LLM} & \textbf{GT} & \textbf{LLM} \\
\midrule
Topic 0 & 0.591 & 0.717 & 2.78 & 3.93 \\
Topic 1 & 0.609 & 0.899 & 3.50 & 3.86 \\
Topic 2 & 0.832 & 0.930 & 3.86 & 3.57 \\
\midrule
\textbf{Overall} & 0.683 & 0.857 & 3.45 & 3.73 \\
\bottomrule
\end{tabular}
\caption{Results for semantic grouping and KP quality.}
\label{tab:grouping_results}
\end{table}

\noindent\textbf{Results.} Table~\ref{tab:grouping_results} shows the results for semantic grouping and KP quality. Ground-truth annotations yield a global cluster precision of 0.683,  indicating that many  arguments assigned to the same KP are not judged to express the same underlying reasoning. In contrast, a non-optimized LLM achieves a cluster precision of 0.857, suggesting that considerably more coherent semantic groupings are possible and that  the semantic groups induced by the original annotations are often inconsistent with the underlying argument structure.

 KP quality shows a similar pattern: ground-truth annotations achieve an average abstraction score of 3.45 compared to 3.73 for the non-optimized LLM, suggesting that many reference KPs do not optimally capture the shared meaning of their assigned arguments and that higher-quality abstractions are achievable without extensive optimization.

\subsection{Evaluation of Redundancy}

\noindent\textbf{Setup.} The evaluation is conducted on the same ArgKP21 test set.
 For each annotation instance, annotators see the topic, stance, and the full set of KPs associated with that topic and stance.
 Annotators are asked to identify the minimum subset of KPs that captures all distinct ideas expressed in the original set. When multiple KPs express the same underlying reasoning, annotators retain only the KP that they judge to be the clearest, most complete, and most representative formulation of that idea. KPs are considered redundant if they express the same underlying argument or claim regardless of phrasing,  and are retained when they reflect distinct reasons or perspectives for the topic.
 Annotators also  provide a brief justification explaining which KPs were judged semantically equivalent and why.

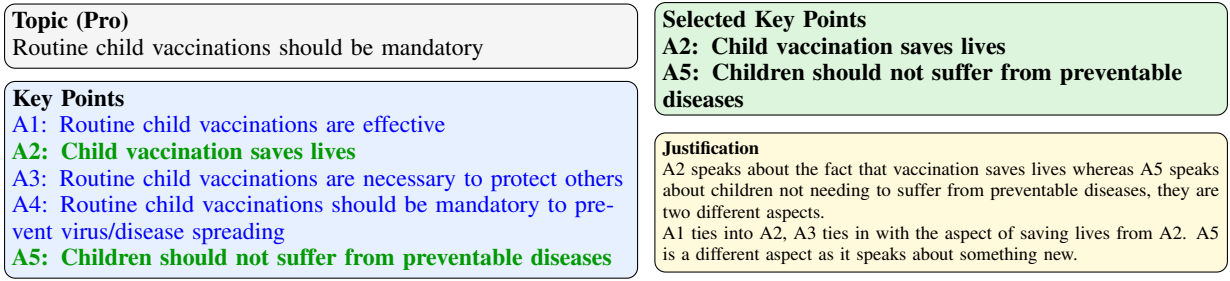
\begin{figure*}[t]
\centering
\begin{tikzpicture}[
    box/.style={
        draw, rounded corners, align=left, inner sep=3pt,
        text width=0.46\textwidth, font=\small
    }
]

 \definecolor{topiccolor}{RGB}{245,245,245}
\definecolor{kpcolor}{RGB}{230,240,255}
\definecolor{selectcolor}{RGB}{220,245,220}
\definecolor{justcolor}{RGB}{255,250,220}

\node[box, text width=0.51\textwidth, fill=topiccolor] (topic) at (0,0) {
    \textbf{Topic (Pro)} \\
    Routine child vaccinations should be mandatory
};

\node[box, text width=0.51\textwidth, fill=kpcolor] (kp) at (0,-1.9) {
    \textbf{Key Points} \\
    {\color{blue} A1: Routine child vaccinations are effective} \\
    {\color{green!60!black} \textbf{A2: Child vaccination saves lives}} \\
    {\color{blue} A3: Routine child vaccinations are necessary to protect others} \\
    {\color{blue} A4: Routine child vaccinations should be mandatory to prevent virus/disease spreading} \\
    {\color{green!60!black} \textbf{A5: Children should not suffer from preventable diseases}}
};

\node[box, text width=0.46\textwidth, fill=selectcolor] (selected) at (8.2,-0.3) {
    \textbf{Selected Key Points} \\
    \textbf{A2: Child vaccination saves lives} \\
    \textbf{A5: Children should not suffer from preventable diseases}
};

\node[box, text width=0.46\textwidth, font=\scriptsize, fill=justcolor] (justification) at (8.2,-2.2) {
    \textbf{Justification} \\
    A2 speaks about the fact that vaccination saves lives whereas A5 speaks about children not needing to suffer from preventable diseases, they are two different aspects. \\
    A1 ties into A2, A3 ties in with the aspect of saving lives from A2. A5 is a different aspect as it speaks about something new.
};

\end{tikzpicture}

\caption{Example of redundancy annotation. KPs A1, A3, and A4 (blue) express similar reasoning and are consolidated under A2 (``saving lives''), while A5 represents a distinct semantic aspect.}
\label{fig:redundancy_example}
\end{figure*}

\noindent\textbf{Metrics.} We quantify redundancy by measuring the proportion of KPs that represent distinct semantic ideas after deduplication.
 For each annotation instance $i$, we define
\(
\text{Uniqueness}(i) = \frac{|\mathcal{S}_i|}{|\mathcal{K}_i|}
\)
where $\mathcal{K}_i$ is the original set of KPs and $\mathcal{S}_i$ is the subset selected by annotators as representing distinct ideas.
A lower uniqueness score indicates higher redundancy, as more KPs are judged to express  overlapping or equivalent meanings.

\noindent\textbf{Results.} The ground-truth annotations achieve a micro-averaged uniqueness score of 0.525, indicating that nearly half of the annotated KPs can be removed without losing distinct ideas. This reveals substantial redundancy in the reference annotations. In contrast, a non-optimized LLM reaches 0.952, suggesting that far more compact and non-redundant representations are achievable.

    Figure~\ref{fig:redundancy_example} shows an example where several KPs describing
 effectiveness, protecting others, and preventing disease spread are grouped under a single core idea (``saving lives''), while ``preventing suffering'' is identified as a distinct semantic dimension.
 This example also reveals a deeper structural issue: KPs are defined at inconsistent levels of abstraction,
mixing  high-level outcomes (e.g. ``saving lives'')
with  lower-level mechanisms (e.g. preventing disease spread or protecting others),
creating implicit subsumption relationships.   Such abstraction-level inconsistency violates the assumption that KPs form a flat set of distinct ideas, instead, revealing  a hierarchical structure in which multiple lower-level KPs correspond to a single higher-level concept.
This contributes to redundancy and prevents the KP set from providing a compact and coherent representation of the argument space.

Overall,  redundancy in the ground-truth annotations is both quantitative and structural: the dataset contains numerous  overlapping KPs and lacks  a consistent level of semantic abstraction, resulting in fragmented representations of underlying ideas.

\subsection{Evaluation of Unmatched Arguments}

\noindent\textbf{Setup.} ArgKP21 provides argument-KP mappings in which some arguments are labeled as unmatched, indicating that they are not covered by any annotated KP.
To assess the validity of these labels, we extract all unmatched arguments in the ground-truth annotations. Due to the large number of unmatched arguments, we focus on the first topic in the test set.
 To make annotation manageable, the set is divided into two subsets, each annotated independently by a different annotator.

For each annotation instance, annotators are shown the topic, a single argument labeled as unmatched, and the full
KP set for that  topic and stance. They are asked to determine if the argument matches  any of the provided KPs by selecting the KP that expresses the same underlying reasoning as the argument, or \textit{None} if no KP applies.

\noindent\textbf{Metrics.} We evaluate unmatched arguments using
\(
\text{True Unmatched Rate} = \frac{|\mathcal{U}_{true}|}{|\mathcal{U}|}
\)
where $\mathcal{U}$ denotes the set of arguments labeled as unmatched in the ground-truth annotations and $\mathcal{U}_{true}$ denotes the subset of these arguments that annotators also classify as unmatched.
 This metric captures the validity of unmatched labels: lower values indicate that many arguments are incorrectly labeled as unmatched.

\noindent\textbf{Results.} The overall true unmatched rate is 0.336, indicating that only about a third of arguments labeled unmatched are confirmed as such under human evaluation.
This implies that most unmatched arguments  can, in fact, be associated with existing KPs, revealing substantial inconsistencies in the original argument-KP mappings.
 Results are consistent across both subsets, with true unmatched rates of 0.400 and 0.281, respectively.
Both values indicate significant disagreement with the ground-truth annotations, showing that the issue is not specific to a particular subset or annotator. At the same time, a non-trivial proportion of arguments remains unmatched even after human evaluation.
These cases indicate that the annotated KP set does not fully cover the argument space, as some arguments express ideas that are not represented by any KP.
 These findings reveal two complementary limitations of the ground-truth annotations: mislabeling of matchable arguments as unmatched  and incomplete coverage of the argument space.

\noindent\textbf{Overall Summary.} Across all three evaluations, ArgKP21 ground-truth annotations exhibit systematic structural limitations. Semantic grouping analysis reveals inconsistent argument–KP assignments, leading to low cluster precision. Redundancy analysis identifies overlapping and hierarchically related KPs rather than distinct KPs.  The unmatched arguments analysis uncovers both incorrect labeling and incomplete coverage of the argument space. Together, these findings show that the annotations fail to provide a coherent, minimal, and semantically faithful representation of the underlying arguments.

These limitations have direct implications for evaluation. Since the ground-truth annotations are not optimal, they cannot serve as an upper bound on performance, resulting in a \textbf{ceiling violation}. Moreover, because standard evaluation methods assume that outputs closer  to the ground truth are better, flawed annotations lead to \textbf{selection failure}, systematically undervaluing higher-quality KP sets.  This challenges the validity of reference-based evaluation for KPA and motivates improved annotation and evaluation frameworks.

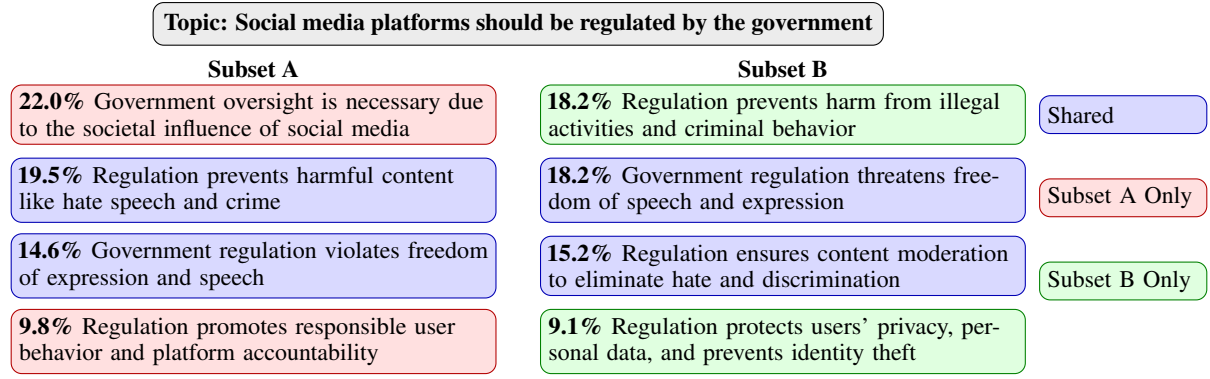
\begin{figure*}[t]
\centering
\footnotesize
\begin{tikzpicture}[
    shared/.style={
        rectangle,
        rounded corners,
        draw=blue!70!black,
        fill=blue!15,
        text width=6.2cm,
        align=left,
        font=\footnotesize
    },
    uniqueA/.style={
        rectangle,
        rounded corners,
        draw=red!70!black,
        fill=red!12,
        text width=6.2cm,
        align=left,
        font=\small
    },
    uniqueB/.style={
        rectangle,
        rounded corners,
        draw=green!50!black,
        fill=green!12,
        text width=6.2cm,
        align=left,
        font=\small
    },
    title/.style={
        font=\bfseries
    }
]

 \node[
    draw,
    rounded corners,
    fill=gray!15,
    inner sep=4pt
] (topic) at (-2.5,4.8)
{\textbf{Topic: Social media platforms should be regulated by the government}};

 \node[title] at (-6,4.2) {Subset A};

\node[uniqueA] at (-6,3.6)
{\textbf{22.0\%} Government oversight is necessary due to the societal influence of social media};

\node[shared] at (-6,2.6)
{\textbf{19.5\%} Regulation prevents harmful content like hate speech and crime};

\node[shared] at (-6,1.6)
{\textbf{14.6\%} Government regulation violates freedom of expression and speech};

\node[uniqueA] at (-6,0.6)
{\textbf{9.8\%} Regulation promotes responsible user behavior and platform accountability};

 \node[title] at (1,4.2) {Subset B};

\node[uniqueB] at (1,3.6)
{\textbf{18.2\%} Regulation prevents harm from illegal activities and criminal behavior};

\node[shared] at (1,2.6)
{\textbf{18.2\%} Government regulation threatens freedom of speech and expression};

\node[shared] at (1,1.6)
{\textbf{15.2\%} Regulation ensures content moderation to eliminate hate and discrimination};

\node[uniqueB] at (1,0.6)
{\textbf{9.1\%} Regulation protects users' privacy, personal data, and prevents identity theft};

 \node[
    shared,
    text width=2cm,
    minimum height=0.5cm
] at (5.5,3.6) {Shared};

\node[
    uniqueA,
    text width=2cm,
    minimum height=0.5cm
] at (5.5,2.5) {Subset A Only};

\node[
    uniqueB,
    text width=2cm,
    minimum height=0.5cm
] at (5.5,1.4) {Subset B Only};

\end{tikzpicture}

\caption{
Two independently sampled argument subsets from the same topic produce different KPA structures. Blue nodes denote themes shared across both subsets, while red and green nodes represent distribution-specific themes.
 }
\label{fig:distribution_sensitivity}
\end{figure*}

\section{Constructing a Distribution-Sensitive Benchmark for True KPA} \label{section: dataset reannotation}

To address these issues, we construct ArgKP-X, a structure-aware, distribution-sensitive benchmark from the ArgKP21 test set through a human-in-the-loop re-annotation process.\footnote{See Appendix \ref{appendix: human instructions} for the human annotation instructions.}

\subsection{Argument Subset Construction}
To construct a distribution-sensitive evaluation set, we derive multiple argument subsets from the original ArgKP21 test split.
 For each topic, we generate five subsets by randomly sampling 30-50 arguments from the full pool, without replacement within a subset (no duplicates) but with replacement across subsets (allowing overlap between subsets).   Each subset is treated as an independent KPA instance and contains the topic and the set of sampled arguments, yielding 15 instances (3 topics $\times$ 5 subsets). This enables us to examine how variations in the underlying argument distribution influence the resulting KPA structure, allowing us to assess whether different samples from the same topic give rise to different semantic groups, KPs, and prevalence estimates.

\subsection{Annotation Workflow} \label{section: annotation workflow}
We construct refined annotations for each argument instance through a controlled human-in-the-loop workflow initialized by LLM outputs.

\noindent\textbf{Initial Draft Generation.}
    For each instance, a strong LLM\footnote{We use Qwen3 235B A22B Instruct 2507.} generates an initial KPA structure consisting of semantic argument groups and their corresponding KPs.\footnote{See Appendix \ref{appendix: llm prompts} for the prompt.} The resulting argument-KP assignments serve as draft annotations for subsequent human refinement.

 \noindent\textbf{Stage 1: Semantic Grouping and KP verification.}
Annotators review each candidate KP together with its assigned arguments and retain only the  arguments that genuinely express the same underlying reasoning, removing those that are inconsistent with the KP, loosely related, or incorrectly assigned. This stage focuses on improving the precision of argument-KP alignment by eliminating erroneous groupings from the initial generation.
\noindent\textbf{Stage 2: Argument Reassignment.}
In the second stage, annotators reconsider all arguments that were not assigned to any KP in Stage~1. For each such argument, annotators decide whether it can be assigned to an existing KP based on semantic compatibility or it does not match any KP and should remain unassigned.

\noindent\textbf{Justification Collection.}
 Throughout the annotation process, annotators provide brief justifications for their decisions.
 These explanations serve as a quality control mechanism and are released as an additional resource for future research.

\noindent\textbf{Dataset Finalization.}
After the two annotation stages,  all arguments assigned to KPs in either Stage~1 or Stage~2 are merged to obtain the final argument-KP mappings. Arguments still unmatched after Stage~2 are manually reviewed by the first author
and each argument is:
(i) assigned to an existing KP if semantically consistent with an existing group,
 (ii) used to create a new KP if it expresses
a recurring uncovered  reasoning pattern, or (iii) left  unmatched if it expresses an isolated, vague, off-topic, or otherwise non-representative idea that cannot be reliably assigned to the discovered KP structure.
 Finally,  KPs with no supporting arguments  are removed from the final dataset.

\noindent\textbf{Final Re-annotated Dataset.}
 The resulting dataset consists of a set of KPA instances, each defined by a topic and a set of semantic argument groups. Each group contains arguments expressing a shared underlying idea and is represented by a corresponding KP. Combined with the use of multiple subsets per topic, the benchmark supports distribution-sensitive analysis of semantic grouping, KP generation, coverage, and prevalence.

\subsection{Illustrating Distribution Sensitivity}

Figure~\ref{fig:distribution_sensitivity} shows two independently sampled argument subsets from the topic \textit{``Social media platforms should be regulated by the government''}. Although both subsets contain common themes such as freedom of expression and harmful-content regulation, they yield different KPA structures. Subset~A emphasizes societal influence and platform accountability, whereas Subset~B focuses on crime prevention and user data protection. This example illustrates that KPA structures are shaped not only by the topic but also by the underlying argument distribution. Thus, different samples of arguments can produce different KPs and prevalence distributions even when the topic remains unchanged.

\section{Benchmark Validation}

To assess whether the proposed benchmark better captures KPA's structural requirements, we compare structures derived from the original ArgKP21 annotations and our reannotated benchmark on the same argument subsets.

\subsection{Evaluation Setup}

 As discussed in Section \ref{section: true kpa},
 a benchmark that faithfully evaluates KPA should capture the underlying structure induced by a given argument collection. For each argument subset in our benchmark, we construct two candidate structures: one derived from the original ArgKP21 annotations by retaining the subset arguments and their corresponding argument–KP mappings, and one from our re-annotated benchmark. Since both structures are evaluated on the same argument subset, any observed differences reflect how the two annotation schemes recover the semantic organization of the argument collection rather than differences in the underlying arguments.

\subsection{LLM Evaluation}

We conduct an automatic evaluation using LLMs.\footnote{See Appendix \ref{appendix: llm prompts} for the prompt.}
For each argument subset, the evaluator receives the topic, the full set of arguments, and two candidate structures (one from the original ArgKP21 annotations and one from our re-annotation procedure),
 and is instructed to assess each structure independently according to the four dimensions of true KPA:
\textbf{semantic grouping},
 \textbf{KP generation},
 \textbf{coverage},
 \textbf{prevalence}.
 Each dimension is scored on a five-point scale from 1 (Very Poor) to 5 (Excellent), and then summed to obtain the final score.   The evaluator additionally selects the structure that better satisfies the requirements of true KPA.
Structure order is randomized to reduce position bias.

\begin{table}[t]
\centering
\small
\begin{tabular}{lcc}
\toprule
Judge & ArgKP21 & Reannotated \\
\midrule
GLM 5 & 8.53 & \textbf{19.80} \\
DeepSeek V3.2 & 9.53 & \textbf{19.40} \\
\bottomrule
\end{tabular}
\caption{Average overall scores across all five subsets.}
\label{tab:llm_eval}
\end{table}

\noindent\textbf{Results.} Table~\ref{tab:llm_eval} shows the average overall scores obtained by two independent LLM judges, GLM 5 and DeepSeek V3.2.\footnote{Detailed dimension-level results and per-subset evaluations are provided in Appendix \ref{appendix: llm eval results}.} Both judges consistently preferred the re-annotated structures over the original ArgKP21 annotations across all five subsets in all topics, yielding in 100\% agreement.  The re-annotated structures substantially outperform the original annotations on all four dimensions of true KPA, suggesting that the proposed benchmark produces more coherent semantic groupings, more representative KPs, more comprehensive coverage, and prevalence estimates that better reflect the underlying argument distribution. Consistency across evaluators further suggests that these improvements are robust to the choice of LLM judge.

\subsection{Human Evaluation}

We sample 15 representative arguments per instance  using embedding-based diversity sampling: arguments were encoded with a sentence transformer\footnote{all-mpnet-base-v2} and grouped into eight semantic clusters\footnote{Both our re-annotations and the original annotations yield around 8-10 KPs per topic.} using K-Means. We select one representative argument closest to each cluster centroid and supplement with random samples  from the remaining pool, covering  major semantic themes while keeping human annotation costs manageable.
  For both structures, only the sampled arguments were retained, and all mappings were restricted to this subset of arguments.
To avoid presentation bias, structure order was randomized per instance.  Annotators were shown two anonymized structures  without knowing which was the original.  Annotators independently evaluated each structure on three dimensions: \textbf{semantic grouping},
 \textbf{KP quality} and
 \textbf{coverage}.
  using a five-point scale from 1 (\textit{Very Poor}) to 5 (\textit{Excellent}),
then  selected the structure that provided the better overall representation of the argument collection providing a written justification for their choice.  Each annotator evaluated one subset from each of the three topics.

\noindent\textbf{Results.} Table~\ref{tab:human_eval_main} summarizes the overall results, while detailed subset-level results are reported in Appendix~\ref{app:human_results}. Across all three dimensions, annotators consistently scored the re-annotated structures substantially higher than the structures derived from the original ArgKP21 annotations.
  Furthermore, annotators preferred the re-annotated structure in all 15 instances, yielding an overall preference rate of 100\%,  regardless of presentation order, suggesting that the observed improvements cannot be explained by position bias.
These findings are consistent with the LLM-based evaluation and independently support our claim that the re-annotated benchmark provides more coherent semantic groupings, higher-quality KPs, and better coverage than  the original ArgKP21 annotations.

\begin{table}[t]
\centering
\small
\begin{tabular}{lp{1.2cm}p{1.5cm}p{1cm}}
\toprule
\textbf{Structure} & \textbf{Grouping} & \textbf{KP Quality} & \textbf{Coverage} \\
\midrule
ArgKP21-derived & 3.00 & 3.07 & 2.60 \\
Reannotated & 4.73 & 4.60 & 4.80 \\
  \bottomrule
\end{tabular}
\caption{Average human evaluation scores across all subsets.}
\label{tab:human_eval_main}
\end{table}

\noindent\textbf{Inter-Annotator Agreement.} To assess reliability, two subsets in each of the three topics were independently evaluated by two annotators, resulting in six
 evaluation instances (2 subsets $\times$ 3 topics). Agreement on overall structure preference was 100\% (6/6 instances), with both annotators consistently preferring the re-annotated structures. Given the ordinal five-point ratings, we report quadratic weighted Cohen's $\kappa$. Agreement was substantial for semantic grouping ($\kappa=0.661$) and KP quality ($\kappa=0.766$), and almost perfect for coverage ($\kappa=0.878$), with exact agreement rates of 66.7\%, 58.3\%, and 58.3\% respectively. These results indicate that the preference for re-annotated structures is reliable and robust across annotators.

\section{Additional Dataset Contributions}

Beyond the distribution-sensitive benchmark, we release several additional datasets produced during benchmark construction and validation.\footnote{Detailed annotation schemas are in Appendix~\ref{app:data_formats}.}

\noindent\textbf{Semantic Group Verification Annotations.}
During benchmark construction, annotators reviewed generated argument groups, identified arguments that belonged to the target semantic group, assessed KP quality, and provided justifications. These annotations support research on semantic grouping, coverage evaluation, and explainable KPA.

\noindent\textbf{Unmatched Argument Reassignment Annotations.}
For arguments initially labeled as unmatched, annotators either assigned them to an existing KP or confirmed that none applied, with accompanying justifications, providing a resource for argument-KP matching and explainable KPA.

\noindent\textbf{Human and LLM Evaluation Annotations.}
Benchmark validation additionally produced human and LLM evaluation datasets containing quality assessments, structure preferences, and explanations. These resources support future work on KPA evaluation, preference prediction, human-LLM agreement, and LLM-as-a-judge methods.

\section{A Research Agenda for True KPA}
Viewing KPA as a structured prediction problem opens several research directions beyond KP generation and mapping. First, future work may investigate \textbf{structure recovery} by jointly identifying semantic argument groups, generating representative KPs, estimating prevalence and coverage rather than treating these independently. Second, our \textbf{distribution-sensitive} benchmark shows that KPA structures depend on the underlying argument distribution. Different samples from the same topic can yield different KPs and prevalence estimates, motivating the study of  how sensitive KPA outputs are to changes in the underlying argument distribution and how prevalence estimates can be robustly measured.
 Third, our formulation supports \textbf{explainable KPA}, grounding KPs explicitly in semantic groups of supporting arguments and enabling  transparent justifications and evidence for generated KPs. Finally, our findings motivate the development of \textbf{structure-aware evaluation methods} that assess semantic grouping, coverage, and prevalence in addition to the quality of generated KPs.
We hope this formulation of true KPA  encourages future work to move beyond KP generation and mapping and toward recovering the underlying organization of argument spaces.

\section{Conclusion}
We revisit KPA  as a structured prediction problem.
 Under this formulation, we show that existing KPA benchmarks exhibit systematic limitations,
 causing ceiling violation and selection failure in reference-based evaluation.  We introduce a structure-aware, distribution-sensitive benchmark built via human-in-the-loop re-annotation. Human and LLM evaluations show that it yields more coherent argument groupings, higher-quality KPs, better coverage, and more reliable prevalence estimates than existing annotations. We release the benchmark and accompanying annotation resources to support future research on KPA evaluation, argument clustering, argument–KP matching, and explainable KPA.
\section{Limitations}

First, although our analyses and validation experiments consistently support our reannotated benchmark's ability to better capture the structural requirements of true KPA, future extensions to additional domains and argument sources may further increase the scope and utility of the benchmark.

Second, while we release a re-annotated benchmark for evaluation, we do not construct a corresponding training set. As a result, the benchmark is primarily intended for assessing KPA systems rather than supporting supervised training. Extending the proposed annotation framework to larger-scale training data may facilitate the development of models that explicitly recover the semantic structures required for true KPA.

Finally, our work focuses on task formulation, dataset diagnosis, benchmark construction, and benchmark validation. We validate our claims through systematic human and LLM-based evaluations, demonstrating that the proposed benchmark better captures the structural properties required for true KPA. We leave large-scale evaluation of existing and future KPA systems under the proposed framework to future work.

   \bibliography{custom}

@inproceedings{bar-haim-etal-2020-arguments,
    title = "From Arguments to Key Points: {T}owards Automatic Argument Summarization",
    author = "Bar-Haim, Roy  and
      Eden, Lilach  and
      Friedman, Roni  and
      Kantor, Yoav  and
      Lahav, Dan  and
      Slonim, Noam",
    editor = "Jurafsky, Dan  and
      Chai, Joyce  and
      Schluter, Natalie  and
      Tetreault, Joel",
    booktitle = "Proceedings of the 58th Annual Meeting of the Association for Computational Linguistics",
    month = jul,
    year = "2020",
    address = "Online",
    publisher = "Association for Computational Linguistics",
    url = "https://aclanthology.org/2020.acl-main.371/",
    doi = "10.18653/v1/2020.acl-main.371",
    pages = "4029--4039"
}

@inproceedings{friedman-etal-2021-overview,
    title = "Overview of the 2021 Key Point Analysis Shared Task",
    author = "Friedman, Roni  and
      Dankin, Lena  and
      Hou, Yufang  and
      Aharonov, Ranit  and
      Katz, Yoav  and
      Slonim, Noam",
    editor = "Al-Khatib, Khalid  and
      Hou, Yufang  and
      Stede, Manfred",
    booktitle = "Proceedings of the 8th Workshop on Argument Mining",
    month = nov,
    year = "2021",
    address = "Punta Cana, Dominican Republic",
    publisher = "Association for Computational Linguistics",
    url = "https://aclanthology.org/2021.argmining-1.16/",
    doi = "10.18653/v1/2021.argmining-1.16",
    pages = "154--164"
}

@inproceedings{gurjar-etal-2025-argcmv,
    title = "{A}rg{CMV}: An Argument Summarization Benchmark for the {LLM}-era",
    author = "Gurjar, Omkar  and
      Goyal, Agam  and
      Chandrasekharan, Eshwar",
    editor = "Christodoulopoulos, Christos  and
      Chakraborty, Tanmoy  and
      Rose, Carolyn  and
      Peng, Violet",
    booktitle = "Proceedings of the 2025 Conference on Empirical Methods in Natural Language Processing",
    month = nov,
    year = "2025",
    address = "Suzhou, China",
    publisher = "Association for Computational Linguistics",
    url = "https://aclanthology.org/2025.emnlp-main.1110/",
    doi = "10.18653/v1/2025.emnlp-main.1110",
    pages = "21870--21883",
    ISBN = "979-8-89176-332-6"
}

@inproceedings{li-etal-2023-hear,
    title = "Do You Hear The People Sing? Key Point Analysis via Iterative Clustering and Abstractive Summarisation",
    author = "Li, Hao  and
      Schlegel, Viktor  and
      Batista-Navarro, Riza  and
      Nenadic, Goran",
    editor = "Rogers, Anna  and
      Boyd-Graber, Jordan  and
      Okazaki, Naoaki",
    booktitle = "Proceedings of the 61st Annual Meeting of the Association for Computational Linguistics (Volume 1: Long Papers)",
    month = jul,
    year = "2023",
    address = "Toronto, Canada",
    publisher = "Association for Computational Linguistics",
    url = "https://aclanthology.org/2023.acl-long.786/",
    doi = "10.18653/v1/2023.acl-long.786",
    pages = "14064--14080"
}

@inproceedings{van-der-meer-etal-2024-empirical,
    title = "An Empirical Analysis of Diversity in Argument Summarization",
    author = "van der Meer, Michiel  and
      Vossen, Piek  and
      Jonker, Catholijn M.  and
      Murukannaiah, Pradeep K.",
    editor = "Graham, Yvette  and
      Purver, Matthew",
    booktitle = "Proceedings of the 18th Conference of the European Chapter of the Association for Computational Linguistics (Volume 1: Long Papers)",
    month = mar,
    year = "2024",
    address = "St. Julian{'}s, Malta",
    publisher = "Association for Computational Linguistics",
    url = "https://aclanthology.org/2024.eacl-long.123/",
    doi = "10.18653/v1/2024.eacl-long.123",
    pages = "2028--2045"
}

@inproceedings{khosravani-etal-2024-enhancing,
    title = "Enhancing Argument Summarization: Prioritizing Exhaustiveness in Key Point Generation and Introducing an Automatic Coverage Evaluation Metric",
    author = "Khosravani, Mohammad  and
      Huang, Chenyang  and
      Trabelsi, Amine",
    editor = "Duh, Kevin  and
      Gomez, Helena  and
      Bethard, Steven",
    booktitle = "Proceedings of the 2024 Conference of the North American Chapter of the Association for Computational Linguistics: Human Language Technologies (Volume 1: Long Papers)",
    month = jun,
    year = "2024",
    address = "Mexico City, Mexico",
    publisher = "Association for Computational Linguistics",
    url = "https://aclanthology.org/2024.naacl-long.454/",
    doi = "10.18653/v1/2024.naacl-long.454",
    pages = "8212--8224"
}

@inproceedings{altemeyer2025argument,
  title={Argument summarization and its evaluation in the era of large language models},
  author={Altemeyer, Moritz and Eger, Steffen and Daxenberger, Johannes and Chen, Yanran and Altendorf, Tim and Cimiano, Philipp and Schiller, Benjamin},
  booktitle={Proceedings of the 2025 Conference on Empirical Methods in Natural Language Processing},
  pages={35478--35499},
  year={2025}
}

@inproceedings{chen-etal-2019-seeing,
    title = "Seeing Things from a Different Angle:Discovering Diverse Perspectives about Claims",
    author = "Chen, Sihao  and
      Khashabi, Daniel  and
      Yin, Wenpeng  and
      Callison-Burch, Chris  and
      Roth, Dan",
    editor = "Burstein, Jill  and
      Doran, Christy  and
      Solorio, Thamar",
    booktitle = "Proceedings of the 2019 Conference of the North {A}merican Chapter of the Association for Computational Linguistics: Human Language Technologies, Volume 1 (Long and Short Papers)",
    month = jun,
    year = "2019",
    address = "Minneapolis, Minnesota",
    publisher = "Association for Computational Linguistics",
    url = "https://aclanthology.org/N19-1053/",
    doi = "10.18653/v1/N19-1053",
    pages = "542--557"
}

@inproceedings{alshomary-etal-2021-key,
    title = "Key Point Analysis via Contrastive Learning and Extractive Argument Summarization",
    author = {Alshomary, Milad  and
      Gurcke, Timon  and
      Syed, Shahbaz  and
      Heinisch, Philipp  and
      Splieth{\"o}ver, Maximilian  and
      Cimiano, Philipp  and
      Potthast, Martin  and
      Wachsmuth, Henning},
    editor = "Al-Khatib, Khalid  and
      Hou, Yufang  and
      Stede, Manfred",
    booktitle = "Proceedings of the 8th Workshop on Argument Mining",
    month = nov,
    year = "2021",
    address = "Punta Cana, Dominican Republic",
    publisher = "Association for Computational Linguistics",
    url = "https://aclanthology.org/2021.argmining-1.19/",
    doi = "10.18653/v1/2021.argmining-1.19",
    pages = "184--189"
}

@inproceedings{kapadnis-etal-2021-team,
    title = "Team Enigma at {A}rg{M}ining-{EMNLP} 2021: Leveraging Pre-trained Language Models for Key Point Matching",
    author = "Kapadnis, Manav  and
      Patnaik, Sohan  and
      Panigrahi, Siba  and
      Madhavan, Varun  and
      Nandy, Abhilash",
    editor = "Al-Khatib, Khalid  and
      Hou, Yufang  and
      Stede, Manfred",
    booktitle = "Proceedings of the 8th Workshop on Argument Mining",
    month = nov,
    year = "2021",
    address = "Punta Cana, Dominican Republic",
    publisher = "Association for Computational Linguistics",
    url = "https://aclanthology.org/2021.argmining-1.21/",
    doi = "10.18653/v1/2021.argmining-1.21",
    pages = "200--205"
}

@inproceedings{reimer-etal-2021-modern,
    title = "Modern Talking in Key Point Analysis: Key Point Matching using Pretrained Encoders",
    author = "Reimer, Jan Heinrich  and
      Luu, Thi Kim Hanh  and
      Henze, Max  and
      Ajjour, Yamen",
    editor = "Al-Khatib, Khalid  and
      Hou, Yufang  and
      Stede, Manfred",
    booktitle = "Proceedings of the 8th Workshop on Argument Mining",
    month = nov,
    year = "2021",
    address = "Punta Cana, Dominican Republic",
    publisher = "Association for Computational Linguistics",
    url = "https://aclanthology.org/2021.argmining-1.18/",
    doi = "10.18653/v1/2021.argmining-1.18",
    pages = "175--183"
}

@inproceedings{phan-etal-2021-matching,
    title = "Matching The Statements: A Simple and Accurate Model for Key Point Analysis",
    author = "Phan, Viet Hoang  and
      Nguyen, Tien Long  and
      Nguyen, Duc Long  and
      Doan, Ngoc Khanh",
    editor = "Al-Khatib, Khalid  and
      Hou, Yufang  and
      Stede, Manfred",
    booktitle = "Proceedings of the 8th Workshop on Argument Mining",
    month = nov,
    year = "2021",
    address = "Punta Cana, Dominican Republic",
    publisher = "Association for Computational Linguistics",
    url = "https://aclanthology.org/2021.argmining-1.17/",
    doi = "10.18653/v1/2021.argmining-1.17",
    pages = "165--174"
}

\appendix

\section{Dataset Limitations} \label{appendix: dataset limitations}

Given the requirements of true KPA, we examine whether existing datasets can evaluate true KPA.
 We analyze two categories of datasets: those specifically designed for KPA, and those repurposed for KPA evaluation. While both appear structurally compatible, we show that they exhibit systematic limitations that hinder the evaluation of semantic grouping, abstraction, coverage, and prevalence.

\subsection{Datasets Designed for KPA}

We first analyze datasets explicitly constructed for KPA, including ArgKP \cite{bar-haim-etal-2020-arguments}, ArgKP21 \cite{friedman-etal-2021-overview}, and ArgCMV \cite{gurjar-etal-2025-argcmv}. ArgKP21 extends ArgKP by adding additional test data, while retaining the same annotation framework. These datasets provide arguments, KPs, and annotated mappings between them, and therefore, in principle, contain the structural components required by the task. The mappings can be interpreted as grouping arguments that express similar underlying ideas, while KPs serve as summaries of these groups, and prevalence can be derived from the number of supporting arguments.

For ArgKP and ArgKP21, the primary limitation lies in how KPs and mappings are constructed.
In these datasets, KPs are written independently of the underlying argument sets and reflect general topic-level statements rather than abstractions grounded in the arguments. Furthermore, argument–key point mappings are established through an independent pairwise matching process rather than being derived jointly from the complete argument set. As a result, both key point generation and argument assignment are decoupled from the underlying argument distribution. Arguments assigned to the same key point therefore do not necessarily form semantically coherent groups, leading to weak alignment between grouping and abstraction. Consequently, the induced grouping does not reflect
 the global organization of the argument corpus, and the natural structure of the argument space.

In addition, these datasets exhibit incomplete coverage of the argument corpus. Using ArgKP21 as a representative example, we find that coverage, defined as the proportion of arguments assigned to at least one key point, is approximately 90-91\%, indicating that around 9-10\% of arguments remain unrepresented. This implies that a non-trivial portion of the argument space is excluded from the final representation, limiting the ability to evaluate whether a system captures the full diversity of opinions.

We further analyze the structure of the mappings. The average number of KPs matched per argument is approximately 1.3-1.4, with about 15\% of arguments associated with multiple KPs. While this reflects the multi-faceted nature of arguments, it also suggests that group boundaries are not clearly defined. Moreover, the distribution of arguments across KPs is highly skewed: a small number of KPs dominate the representation, while others receive only limited support. For instance, in one topic, the most frequent key point is associated with 38 arguments, whereas several KPs are supported by fewer than 10. This long-tailed distribution indicates that the representation is uneven and may fail to capture the full diversity of the argument space.

As a result, prevalence estimates derived from these mappings are sensitive to the underlying annotation process. When KPs do not accurately reflect the underlying argument groups, the resulting distribution becomes unstable and may not reliably represent the true prominence of ideas in the corpus.

In contrast, ArgCMV follows a different construction paradigm. KPs are generated using an LLM conditioned on the topic and the full set of arguments, separately for pro and con stances. While this ensures that KPs are grounded in the data, the generation is performed at the level of the entire argument set rather than individual semantic groups. As a result, argument grouping is not explicitly recovered, and mappings are established only after key point generation. This decouples abstraction from grouping, which contradicts the requirement that KPs should be derived from coherent argument clusters.

Moreover, ArgCMV introduces additional limitations. All annotations are generated by an LLM without human correction, and although human evaluation identifies errors in these annotations, no corrective steps are taken. This raises concerns about annotation reliability. Additionally, the dataset is distributed only as Reddit IDs, and recent platform restrictions make it difficult to access the underlying content, limiting reproducibility and practical usability.

Taken together, these observations show that, although datasets specifically designed for KPA provide the necessary structural elements, they fail to enforce the requirements of true KPA in different ways. ArgKP and ArgKP21 suffer from weak alignment between KPs and argument groups, incomplete coverage, and unstable prevalence, while ArgCMV fails to explicitly recover the underlying grouping structure and introduces additional concerns related to annotation quality and accessibility. Therefore, none of these datasets reliably capture the structured representation required for evaluating true KPA.

\subsection{Datasets Used for KPA Evaluation}
We next consider datasets that are used to evaluate KPA systems. We examine Perspectrum \cite{chen-etal-2019-seeing} as a representative example, as it has been adopted in \citet{van-der-meer-etal-2024-empirical} to assess diversity and coverage in KPA settings. Perspectrum provides claims, perspectives, and evidence statements, which can be interpreted as topics, KPs, and argument–key point mappings \cite{chen-etal-2019-seeing}, forming a hierarchical structure in which arguments are linked to KPs.

However, several characteristics of this structure are not well aligned with the objectives of KPA.

First, the set of KPs for each topic is not consistently defined. Some topics contain only a single key point, others include only pro or only con KPs, while some topics contain a large number of KPs. This variability indicates that the KPs are not constructed as a complete and balanced set of representative ideas for a given topic, but instead reflect annotation decisions that are inconsistent across topics. As a result, the dataset does not provide a stable basis for evaluating coverage or completeness.

Second, although arguments are mapped to KPs, the purpose of this mapping differs fundamentally from the objective of KPA. In Perspectrum, mappings capture whether an argument supports a given key point, rather than grouping arguments into clusters that share the same underlying idea. Consequently, KPs function as predefined claims to be supported, rather than as summaries derived from the argument corpus. This leads to a structural mismatch, where arguments are organized as evidence for claims, rather than being grouped and abstracted into representative KPs.

This mismatch is further amplified by the nature of the arguments themselves. Arguments in Perspectrum are often long and multi-faceted, frequently containing multiple claims or lines of reasoning within a single text. As a result, a single argument may support multiple KPs. While such multi-mapping is not inherently problematic, in this context it reflects the fact that arguments are not segmented or constructed with respect to a single underlying idea. This makes it difficult to interpret KPs as concise abstractions of coherent argument groups.

Taken together, these observations indicate that, although Perspectrum provides explicit argument–key point mappings, its structure is not aligned with KPA as a grouping and abstraction task over a well-defined argument corpus. Instead, it is designed to capture support relationships between arguments and predefined claims, which limits its suitability for evaluating whether systems recover the structured representation required for true KPA. Similar limitations apply to other datasets that can be converted into KPA-style formats, which, while structurally compatible, do not capture the grouping, abstraction, and coverage requirements necessary for evaluating true KPA.

\section{Additional LLM evaluation results} \label{appendix: llm eval results}
Table \ref{tab:llm_eval_dimensions} shows the average dimension-level LLM evaluation scores across all subsets. Table \ref{tab:llm_eval_subsets} shows overall LLM evaluation scores for individual subsets.

\begin{table*}[t]
\centering
\small
\begin{tabular}{llccccc}
\toprule
Judge & Structure &
Grouping &
KP Gen. &
Coverage &
Prevalence &
Overall \\
\midrule
\multirow{2}{*}{GLM}
& ArgKP21
& 2.07 & 2.87 & 1.53 & 2.07 & 8.53 \\
& Reannotated
& \textbf{5.00} & \textbf{5.00}
& \textbf{4.80} & \textbf{5.00}
& \textbf{19.80} \\
\midrule
\multirow{2}{*}{DeepSeek V3.2}
& ArgKP21
& 2.67 & 2.73 & 1.93 & 2.20 & 9.53 \\
& Reannotated
& \textbf{4.93} & \textbf{4.93}
& \textbf{4.60} & \textbf{4.93}
& \textbf{19.40} \\
\bottomrule
\end{tabular}
\caption{Average dimension-level LLM evaluation scores across all five subsets.}
\label{tab:llm_eval_dimensions}
\end{table*}

\begin{table*}[h]
\centering
\small
\begin{tabular}{lcccc}
\toprule
Subset &
GLM (ArgKP21) &
GLM (Reannotated) &
DeepSeek (ArgKP21) &
DeepSeek (Reannotated) \\
\midrule
Subset 0 & 8.33 & \textbf{19.67} & 8.67 & \textbf{19.33} \\
Subset 1 & 9.00 & \textbf{19.67} & 9.67 & \textbf{19.67} \\
Subset 2 & 8.67 & \textbf{20.00} & 10.00 & \textbf{18.67} \\
Subset 3 & 8.33 & \textbf{19.67} & 10.00 & \textbf{19.33} \\
Subset 4 & 8.33 & \textbf{20.00} & 9.33 & \textbf{20.00} \\
\bottomrule
\end{tabular}
\caption{Overall LLM evaluation scores for individual subsets. Both judges preferred the reannotated structures on all five subsets.}
\label{tab:llm_eval_subsets}
\end{table*}

\section{Human Evaluation Results}
\label{app:human_results}

Table~\ref{tab:human_eval_detailed} reports detailed human evaluation results for each evaluated subset.

\begin{table*}[t]
\centering
\small
\begin{tabular}{lcccccc}
\toprule
&
\multicolumn{2}{c}{\textbf{Semantic Grouping}} &
\multicolumn{2}{c}{\textbf{KP Quality}} &
\multicolumn{2}{c}{\textbf{Coverage}} \\
\cmidrule(lr){2-3}
\cmidrule(lr){4-5}
\cmidrule(lr){6-7}
\textbf{Subset}
& \textbf{ArgKP21} & \textbf{Reann.}
& \textbf{ArgKP21} & \textbf{Reann.}
& \textbf{ArgKP21} & \textbf{Reann.} \\
\midrule
Subset 0   & 2.67 & 4.67 & 2.67 & 4.33 & 2.00 & 5.00 \\
Subset 0-2 & 3.33 & 4.33 & 2.67 & 4.67 & 2.67 & 5.00 \\
Subset 1   & 3.33 & 4.67 & 3.00 & 4.33 & 2.67 & 5.00 \\
Subset 1-2 & 3.33 & 4.33 & 2.67 & 4.67 & 2.67 & 5.00 \\
Subset 2   & 2.00 & 4.33 & 3.00 & 4.33 & 2.33 & 4.33 \\
Subset 3   & 3.00 & 5.00 & 3.33 & 5.00 & 2.33 & 5.00 \\
Subset 4   & 4.00 & 5.00 & 3.33 & 5.00 & 3.67 & 4.67 \\
\bottomrule
\end{tabular}
\caption{Detailed human evaluation results for each subset.}
\label{tab:human_eval_detailed}
\end{table*}

\section{Benchmark Statistics}

Our benchmark consists of 15 independently annotated subsets across three debate topics. Since key point analysis is performed at the subset level, we report statistics on a per-subset basis. The benchmark contains an average of 39.3 arguments and 8.3 KPs per subset.
The average numbers of pro and con KPs are 4.2 and 4.1 per subset, respectively. On average, fewer than one argument remains unmatched in each subset (0.93 unmatched arguments per subset).

A semantic group consists of a set of semantically related arguments that express a shared viewpoint. Each semantic group is summarized by a key point. The average semantic group size is 4.66 arguments, with group sizes ranging from 1 to 12 arguments.

\begin{table}[t]
\centering
\small
\begin{tabular}{lr}
\toprule
Statistic & Value \\
\midrule
Topics & 3 \\
Subsets & 15 \\
Avg. Arguments per Subset & 39.27 \\
Median Arguments per Subset & 40 \\
Min Arguments per Subset & 32 \\
Max Arguments per Subset & 48 \\
Avg. KPs per Subset & 8.33 \\
Avg. Pro KPs per Subset & 4.20 \\
Avg. Con KPs per Subset & 4.13 \\
Avg. Semantic Group Size & 4.66 \\
Min Semantic Group Size & 1 \\
Max Semantic Group Size & 12 \\
Avg. Unmatched Arguments per Subset & 0.93 \\
\bottomrule
\end{tabular}
\caption{
Summary statistics of the benchmark. Statistics are reported at the subset level, which is the unit of evaluation in our benchmark.
}
\label{tab:benchmark_stats}
\end{table}

\begin{figure}[t]
    \centering
    \includegraphics[width=\linewidth]{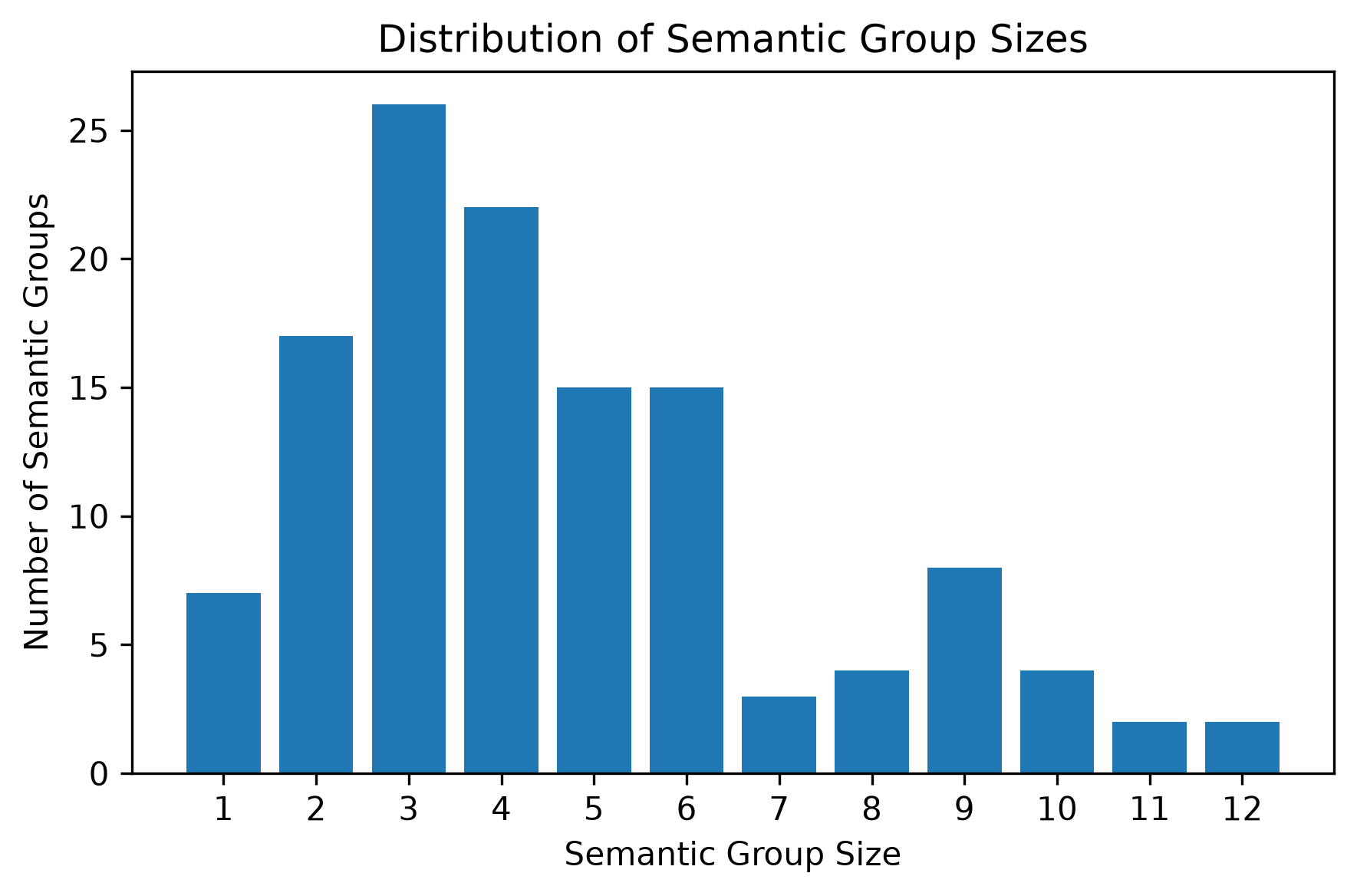}
    \caption{
    Distribution of semantic group sizes in the benchmark.
    A semantic group consists of a set of semantically related arguments summarized by a key point. Most semantic groups contain between three and six arguments, while a small number of larger groups contain up to twelve arguments.
    }
    \label{fig:group_size_distribution}
\end{figure}

Figure~\ref{fig:group_size_distribution} shows the distribution of semantic group sizes. Most semantic groups contain between three and six arguments, indicating that KPs typically summarize multiple semantically related arguments rather than isolated opinions. The distribution is moderately right-skewed, with a small number of larger groups containing up to twelve arguments.

\section{Additional Data Formats and Statistics}
\label{app:data_formats}

\subsection{Semantic Group Verification Annotations}

The semantic group verification dataset records which arguments from a candidate argument group truly belong to a coherent semantic cluster represented by a key point. Annotators additionally assign a key-point quality label and provide a justification.

A typical record contains:

\begin{itemize}
    \item Topic
    \item Key point
    \item Candidate argument group
    \item Selected coherent arguments
    \item Key-point quality label
    \item Annotator justification
\end{itemize}

\paragraph{Data statistics}: Each example consists of a key point, a set of candidate arguments, and human annotations identifying which candidate arguments belong to the semantic group represented by the key point. The dataset contains 126 examples and is balanced across both topics and key-point stances.

Table~\ref{tab:sgv_stats} summarizes the dataset statistics.
Each example contains an average of 4.8 candidate arguments, of which 3.8 arguments are selected as belonging to the semantic group represented by the key point. Candidate set sizes range from 1 to 13 arguments, while the number of selected arguments ranges from 0 to 11.

\begin{table}[t]
\centering
\small
\begin{tabular}{lr}
\toprule
Statistic & Value \\
\midrule
Examples & 126 \\
Topics & 3 \\
Avg. Candidate Arguments per Example & 4.79 \\
Median Candidate Arguments per Example & 4 \\
Min Candidate Arguments per Example & 1 \\
Max Candidate Arguments per Example & 13 \\
Avg. Selected Arguments per Example & 3.81 \\
Median Selected Arguments per Example & 3 \\
Min Selected Arguments per Example & 0 \\
Max Selected Arguments per Example & 11 \\
\bottomrule
\end{tabular}
\caption{
Summary statistics of the semantic group verification dataset.
Selected arguments correspond to candidate arguments judged to belong to the semantic group represented by the key point.
}
\label{tab:sgv_stats}
\end{table}

The dataset is evenly distributed across the three debate topics and contains an equal number of pro and con KPs. The stance distributions are shown in
Tables~\ref{tab:sgv_stance_distribution}.

\begin{table}[t]
\centering
\small
\begin{tabular}{lr}
\toprule
Stance & Examples \\
\midrule
Pro & 63 \\
Con & 63 \\
\bottomrule
\end{tabular}
\caption{
Key-point stance distribution of the semantic group verification
dataset.
}
\label{tab:sgv_stance_distribution}
\end{table}

The dataset also contains the quality of the KPs. Following our annotation guidelines, KPs were rated according to how well they summarize the underlying semantic group of arguments.
A high-quality key point should accurately capture the shared meaning of the selected arguments while remaining concise and representative of the group. Table~\ref{tab:sgv_kp_quality} reports the
distribution of key-point quality labels. Most examples are rated as \textit{Excellent} (84 examples) or \textit{Good}
(37 examples), indicating that the majority of KPs provide high-quality summaries of their corresponding semantic groups.

\begin{table}[t]
\centering
\small
\begin{tabular}{lr}
\toprule
Quality & Count \\
\midrule
Excellent & 84 \\
Good & 37 \\
Fair & 2 \\
Poor & 3 \\
\bottomrule
\end{tabular}
\caption{
Distribution of key-point quality labels in the semantic group verification dataset.
}
\label{tab:sgv_kp_quality}
\end{table}

\subsection{Unmatched Argument Reassignment Annotations}

The unmatched argument dataset records decisions regarding arguments that were initially not assigned to any key point. Annotators either map the argument to an existing key point or determine that it should remain unmatched. Each decision is accompanied by a justification.

Each record contains:

\begin{itemize}
    \item Topic
    \item Unmatched argument
    \item Available KPs
    \item Selected key point (or \texttt{none})
    \item Justification
\end{itemize}

 Summary statistics are reported in Table~\ref{tab:unmatched_reassignment_stats}.

The dataset contains 125 examples spanning three debate topics. Among these examples, 100 (80.0\%) were assigned to at least one existing key point, while 25 (20.0\%) remained unmatched.

\begin{table}[t]
\centering
\small
\begin{tabular}{lr}
\toprule
Statistic & Value \\
\midrule
Examples & 125 \\
Topics & 3 \\
Assigned to at least one key point & 100 \\
No suitable key point (\texttt{none}) & 25 \\
Match Rate (\%) & 80.0 \\
\bottomrule
\end{tabular}
\caption{Statistics of the unmatched argument reassignment dataset.}
\label{tab:unmatched_reassignment_stats}
\end{table}

\subsection{Human Evaluation Annotations}

The human evaluation dataset contains pairwise comparisons between two candidate KPA structures. Annotators assess multiple quality dimensions, select a preferred structure, and provide a justification for their overall preference.

Each record contains:

\begin{itemize}
    \item Topic
    \item Structure A
    \item Structure B
    \item Semantic grouping score
    \item Key-point quality score
    \item Coverage score
    \item Overall preference
    \item Preference justification
\end{itemize}

\subsection{Multi-Judge LLM Evaluation Annotations}

The LLM evaluation dataset contains assessments from two independent LLM judges. Unlike the human evaluation annotations, which provide a single preference-level explanation, the LLM annotations contain explanations for every evaluation dimension in addition to the final comparison rationale.

Each record contains:

\begin{itemize}
    \item Semantic grouping score and reasoning
    \item Key-point generation score and reasoning
    \item Coverage score and reasoning
    \item Prevalence score and reasoning
    \item Final score
    \item Overall preference
    \item Comparison-level justification
\end{itemize}

\section{LLM prompting configuration}
All the LLMs in this paper are accessed via public APIs using the OpenRouter platform (https://openrouter.ai), which provides unified access to multiple model providers. The model and identifier are shown in Table \ref{table: model provider}. We use greedy decoding for all the experiments, \texttt{temperature = 0.0, top\_p = 1.0}.

\begin{table}[t]
\centering
\begin{tabular} {cc}
\hline
\textbf{Model} & \textbf{Identifier} \\
\hline
Qwen 3 235B & \texttt{qwen/qwen3-235b-a22b-2507} \\
GLM 5 & \texttt{z-ai/glm-5} \\
DeepSeek V3.2 & \texttt{deepseek/deepseek-v3.2} \\
\hline
\end{tabular}
\caption{Identifier for the models we use in the prompting experiments.}
\label{table: model provider}
\end{table}

\section{Artifact Documentation and Responsible Use}

\subsection{Artifact Description}

We release a distribution-sensitive benchmark for true Key Point Analysis (KPA), together with several additional annotation resources produced during benchmark construction and validation. These resources include semantic group verification annotations, unmatched argument reassignment annotations, human evaluation annotations, and LLM evaluation annotations.

The benchmark is derived from the ArgKP21 dataset through a human-in-the-loop reannotation process designed to better capture semantic grouping, key point generation, coverage, and prevalence.

\subsection{Intended Use}

The released artifacts are intended for research purposes, including:

\begin{itemize}
    \item Key Point Analysis (KPA)
    \item Argument mining
    \item Argument clustering and grouping
    \item Argument-key point matching
    \item Explainable argument summarization
    \item Evaluation and benchmarking of large language models
\end{itemize}

The benchmark is designed primarily for evaluation and analysis rather than large-scale supervised training.

\subsection{Licensing}

The benchmark and accompanying annotation resources are released under the Apache License 2.0.

The benchmark is derived from ArgKP21. Users must comply with both the license of the original dataset and the license accompanying our released resources. Appropriate attribution to both the original dataset and this work must be preserved in any redistribution or derivative work.

\subsection{Ethical Considerations}

The benchmark is derived from publicly available debate data and may contain subjective, controversial, or conflicting viewpoints originating from the source dataset. These viewpoints do not necessarily reflect the views of the authors.

The annotation process involves both human annotators and large language models. As with any human annotation effort, decisions regarding semantic grouping and abstraction may reflect subjective interpretations of argument meaning. To improve transparency, we release annotation justifications whenever available.

We recommend against using the benchmark as the sole basis for high-stakes decision-making systems without additional validation and human oversight.

\section{Annotator recruitment and payment}

We recruit all the annotators on the Prolific platform. We select annotators who are native English speakers and have passed the AI task annotation test. The annotators are paid 9 pounds per hour, the total cost is around 500 pounds.

\section{LLM prompts} \label{appendix: llm prompts}
\DefineVerbatimEnvironment{prompt}{Verbatim}
  {breaklines=true, breaksymbolleft={}, fontsize=\footnotesize}

Prompt for the LLM generation in Section \ref{section: argkp evaluation}:
\begin{prompt}
You are a professional debater who organises arguments into clear, well-structured reasoning patterns.

====================
Task
====================

You are given:
- A debate topic
- A set of arguments with IDs (each with stance: PRO or CON)

Your goal is to perform high-quality Key Point Analysis (KPA) in a structured way:

1. Identify reasoning-based groups of arguments
2. Ensure each group is semantically coherent
3. Generate a key point that summarises each group
4. Assign arguments to their corresponding key points

====================
Definition (KPA)
====================

A key point:
- expresses ONE underlying reasoning or justification
- summarises a coherent group of arguments
- is a concise and general abstraction of those arguments

CRITICAL PRINCIPLE:
- Arguments belong to the same key point ONLY if they share the SAME reasoning
- Arguments that support the same stance but use DIFFERENT reasons MUST be separated

IMPORTANT:
- Related reasoning is NOT the same as identical reasoning
- Differences in justification, perspective, or logic require separation

====================
STEP 1: Identify Reasoning Groups (Implicit)
====================

- Consider PRO and CON arguments separately
- Identify distinct reasoning patterns
- Group arguments into clusters based on shared reasoning

A valid group:
- contains arguments with the SAME underlying justification
- excludes arguments that differ in reasoning

====================
STEP 2: Control Granularity
====================

- Generate approximately 3-5 key points per stance
- Each key point corresponds to ONE reasoning group

IMPORTANT:
- Do NOT merge different reasoning patterns
- If a group becomes too large:
    → check for sub-reasoning
    → SPLIT into multiple groups if needed

- Large clusters are allowed ONLY if all arguments share exactly the same reasoning

====================
STEP 3: Generate Key Points
====================

For each reasoning group:
- Generate a concise key point describing the shared reasoning

Requirements:
- Must capture the common justification of ALL arguments
- Must be precise and not overly broad
- Must NOT combine multiple distinct reasons
- Must be a single sentence or short phrase

Avoid:
- vague or generic statements
- mixing multiple reasoning types
- restating specific arguments

====================
STEP 4: Assign Arguments
====================

For each argument:
- Assign it to the ONE key point that best matches its reasoning

- Each argument must appear in AT MOST one key point

====================
STEP 5: Coherence and Validation (MANDATORY)
====================

Before producing the output, verify:

For each key point:
1. Do ALL arguments share the SAME reasoning?
2. Are there hidden sub-types of reasoning?
3. Would a human split this into multiple groups?

If YES:
→ SPLIT into multiple key points

Also verify:
- Every assigned argument clearly fits the key point
- No loosely related arguments are included

====================
Output Format (STRICT)
====================

Return a valid JSON object ONLY:

{{
  "pro_keypoints": [
    {{
      "key_point": "<key point>",
      "arguments": ["A1", "A3", "..."]
    }}
  ],
  "con_keypoints": [
    {{
      "key_point": "<key point>",
      "arguments": ["A2", "A5", "..."]
    }}
  ]
}}

====================
Constraints
====================

- Each key point must be less than {kp_token_length} tokens
- Each argument ID may appear in AT MOST one key point
- Use ONLY provided argument IDs
- Do NOT include argument text
- Do NOT include explanations outside JSON

====================
INPUT
====================

Topic:
"{topic}"

Arguments (ID: text):
{arguments}
\end{prompt}

Prompt for the initial KPA structure generation for the benchmark construction in Section \ref{section: dataset reannotation}:
\begin{prompt}
You are a professional debater who organises arguments into clear, well-structured reasoning patterns.

====================
Task
====================

You are given:
- A debate topic
- A set of arguments with IDs (each with stance: PRO or CON)

Your goal is to perform high-quality Key Point Analysis (KPA) by jointly optimising:

1. Semantic Grouping
2. Key Point Generation
3. Coverage
4. Prevalence Representation

Specifically, you must:

1. Identify reasoning-based groups of arguments
2. Ensure each group is semantically coherent
3. Generate a key point that summarises each group
4. Assign arguments to key points
5. Maximise coverage while maintaining coherence
6. Avoid redundant key points

====================
Definition (True KPA)
====================

A key point:
- expresses ONE underlying reasoning or justification
- summarises a coherent group of arguments
- is a concise and general abstraction of those arguments

A correct KPA structure should:
- group semantically similar arguments together
- represent all major reasoning patterns
- minimise redundancy between key points
- maximise coverage of the argument set

CRITICAL PRINCIPLE:
- Arguments belong to the same key point ONLY if they share the SAME core reasoning
- Arguments supporting the same stance but using DIFFERENT reasons MUST be separated

IMPORTANT:
- Related reasoning is NOT necessarily identical reasoning
- Differences in evidence, examples, or wording alone do NOT require a separate key point
- A separate key point is required only when the underlying justification differs

====================
STEP 1: Identify Reasoning Groups
====================

Consider PRO and CON arguments separately.

Identify distinct reasoning patterns and organise arguments into coherent semantic groups.

A valid group:
- contains arguments that share the same core justification
- excludes arguments whose primary reasoning is different
- represents a recurring idea in the argument set

====================
STEP 2: Control Granularity
====================

Generate approximately 3-5 key points per stance.

However:

- Coverage and coherence are more important than a fixed number of key points.
- Create additional key points if necessary to represent distinct recurring reasoning patterns.
- Do not merge different reasoning patterns merely to reduce the number of key points.

Large groups are acceptable only if all arguments genuinely share the same reasoning.

====================
STEP 3: Generate Key Points
====================

For each reasoning group:

Generate a concise key point that represents the shared reasoning.

Requirements:

- Capture the common justification of ALL assigned arguments
- Be precise rather than overly broad
- Represent ONE reasoning pattern
- Be concise and easy to understand

Avoid:

- vague statements
- generic topic summaries
- combining multiple reasons into one key point
- restating individual arguments

====================
STEP 4: Assign Arguments
====================

For each argument:

Assign it to the ONE key point that best represents its primary reasoning.

Assignment rules:

- Focus on the underlying justification rather than wording.
- Minor differences in examples, evidence, or phrasing do not require separation.
- Assign an argument whenever its primary reasoning clearly aligns with a key point.

Each argument may belong to AT MOST one key point.

====================
STEP 5: Coverage Review (MANDATORY)
====================

After completing the initial assignment:

Review ALL unassigned arguments.

For each unassigned argument:

1. Determine whether its reasoning is consistent with an existing key point.

If YES:
→ assign it to that key point.

If NO:

2. Determine whether multiple unassigned arguments express the same recurring reasoning pattern.

If YES:
→ create a new key point and assign them.

Leave an argument unassigned ONLY if:

- it is vague
- it is off-topic
- it expresses an isolated idea that does not form a recurring pattern
- it cannot reasonably be represented by any existing or new key point

Coverage objective:

- Maximise coverage of the argument set.
- Minimise unmatched arguments.
- Prefer creating a coherent new key point over leaving multiple related arguments unmatched.

====================
STEP 6: Coherence, Redundancy and Coverage Validation
====================

Before producing the final output, verify:

For each key point:

1. Do all assigned arguments share the same core reasoning?
2. Are there hidden subgroups with different justifications?
3. Would splitting improve semantic coherence?

If YES:
→ split the key point.

Redundancy check:

1. Do two key points express essentially the same justification?
2. Could they be merged without losing meaning?

If YES:
→ merge them.

Coverage check:

1. Review every unassigned argument again.
2. Can it be assigned to an existing key point?
3. Can it form a new recurring reasoning group together with other unmatched arguments?

If YES:
→ assign it or create a new key point.

Leave arguments unassigned only as a last resort.

Final objective:

- High semantic coherence
- High coverage
- Low redundancy
- Clear prevalence representation through argument assignments

====================
Output Format (STRICT)
====================

Return a valid JSON object ONLY:

{
  "pro_keypoints": [
    {
      "key_point": "<key point>",
      "arguments": ["A1", "A3"]
    }
  ],
  "con_keypoints": [
    {
      "key_point": "<key point>",
      "arguments": ["A2", "A5"]
    }
  ]
}

====================
Constraints
====================

- Each key point must be less than {kp_token_length} tokens
- Each argument ID may appear in AT MOST one key point
- Use ONLY provided argument IDs
- Do NOT include argument text
- Do NOT include explanations outside JSON

Coverage requirement:

- Unassigned arguments should be rare.
- Create new key points when necessary to preserve coverage and coherence.

====================
INPUT
====================

Topic:
"{topic}"

Arguments (ID: text):
{arguments}
\end{prompt}

KPA evaluation prompt:
\begin{prompt}
You are an expert evaluator of Key Point Analysis (KPA).

In this evaluation, KPA is defined as a structured prediction task whose goal is to recover the underlying organization of an argument collection.

A high-quality KPA output should satisfy four requirements:

1. Semantic Grouping
   - Arguments expressing similar ideas should be grouped together.
   - Distinct argument themes should be separated into different groups.
   - The grouping structure should reflect the semantic organization of the argument collection.

2. Key Point Generation
   - Each key point should accurately represent the arguments assigned to it.
   - Key points should capture the shared meaning of the group.
   - Key points should be informative, self-contained, and appropriately abstract.

3. Coverage
   - The collection of key points should provide a complete representation of the major argument themes.
   - Important themes should not be missing.
   - Redundant key points should be avoided.

4. Prevalence
   - The structure should reflect the distribution of arguments.
   - Major argument groups should be appropriately represented.
   - The relative importance of themes should align with the underlying argument collection.

Evaluate Structure A and Structure B independently.

IMPORTANT:

- Do not evaluate similarity to any existing annotation.
- Do not assume either structure is correct.
- Focus on how well each structure satisfies the KPA requirements above.
- Carefully inspect both the key points and their assigned arguments.
- Assign scores independently before determining a winner.

Scoring scale:

1 = Very Poor
2 = Poor
3 = Acceptable
4 = Good
5 = Excellent

The final score for each structure should equal the sum of its four dimension scores.

Return ONLY a valid JSON object using the following schema:

{{
  "structure_a": {{
    "semantic_grouping": {{
      "score": 0,
      "reasoning": ""
    }},
    "key_point_generation": {{
      "score": 0,
      "reasoning": ""
    }},
    "coverage": {{
      "score": 0,
      "reasoning": ""
    }},
    "prevalence": {{
      "score": 0,
      "reasoning": ""
    }},
    "final_score": 0
  }},
  "structure_b": {{
    "semantic_grouping": {{
      "score": 0,
      "reasoning": ""
    }},
    "key_point_generation": {{
      "score": 0,
      "reasoning": ""
    }},
    "coverage": {{
      "score": 0,
      "reasoning": ""
    }},
    "prevalence": {{
      "score": 0,
      "reasoning": ""
    }},
    "final_score": 0
  }},
  "comparison": {{
    "winner": "A",
    "justification": ""
  }}
}}

Valid values for "winner" are:

- "A"
- "B"
- "Tie"

==================================================
INPUT
==================================================

TOPIC

{topic}

==================================================
ARGUMENTS
==================================================

{arguments}

==================================================
CANDIDATE STRUCTURE A
==================================================

{structure_a}

==================================================
CANDIDATE STRUCTURE B
==================================================

{structure_b}
\end{prompt}

\section{Human annotation instructions} \label{appendix: human instructions}
Human annotation consent form:
\begin{prompt}
**Consent Form**

Please read the following information carefully before continuing.

**Purpose of the Study**
In this task, you will evaluate how well a key point represents a set of arguments.
Your responses will be used for research on argument analysis and to improve dataset quality.

**What data we collect:**
- Your corrected labels
- Your Prolific PID
- No personal identifying information is collected

**Voluntary Participation:**
You may withdraw at any time by closing the task window before submission.

**Risks/Benefits:**
Minimal risk. You may stop at any time.

By checking the box below, you confirm you understand the above information and consent to participate.
\end{prompt}

Instructions for semantic grouping and key point verification:
\begin{prompt}
# Annotation Instructions: Key Point Analysis Task

## Goal
In this task, you will evaluate how well a **key point** represents a set of **arguments**.

For each example, you will:
1. Select relevant arguments
2. Explain your decision
3. Rate the quality of the key point

---

## What You Will See
- **Topic** – the general discussion subject
- **Key Point** – a short statement expressing a main idea
- **Arguments** – a list of statements related to the topic

---

## Step 1: Select Arguments

**Select the arguments that express the same main idea as the key point.**

### When selecting arguments:
- Focus on the **meaning and reasoning**, not just shared words
- Include arguments that express the **same underlying reason or claim**
- Exclude arguments that discuss a **different idea** or focus on a different aspect

Use your judgement to decide which arguments best match the key point.

---

## Step 2: Provide a Justification (Required)

Briefly explain **why the selected arguments belong together**.

Your explanation should describe:
- the **shared idea or reasoning** among the selected arguments

### Example:
> “The selected arguments all focus on reducing traffic, while others discuss cost.”

---

## Step 3: Rate Key Point Quality (Required)

**How well does the key point summarise the selected arguments?**

- **Poor** – does not represent the arguments
- **Fair** – partially correct but misses key aspects
- **Good** – mostly accurate with minor issues
- **Excellent** – clearly and accurately summarises the arguments

---

## Example

**Topic:** Transportation
**Key Point:** *“Public transport reduces traffic congestion.”*

**Arguments:**
- A1: “Buses reduce the number of cars on the road”
- A2: “More trains mean fewer commuters driving”
- A3: “Public transport can be expensive”

**Selection:** A1, A2

**Explanation:**
A1 and A2 share the idea of reducing traffic congestion.
A3 discusses cost, which is a different issue.

---

## Important Guidelines
- Focus on **meaning**, not wording
- Different arguments may express the same idea in different ways
- There may be **more than one reasonable selection**

---

## IMPORTANT: Coverage

Your goal is to select ALL arguments that match the key point — no more, no less.

- Do not select only the most obvious or strongest arguments
- Include all arguments that express the SAME reasoning
- If an argument matches the key point, even if phrased differently, it should be selected

- At the same time, do NOT include arguments that express a different reason, even if they are related to the topic

---

### Common Mistakes to Avoid

Annotators often make the following mistakes:

- Selecting only a few “representative” arguments
- Ignoring arguments that are more specific
- Including arguments that are related to the topic but express a different reason

These are incorrect.

---

### General vs Specific Reasoning

Arguments may express the same reasoning at different levels:

- Some are general
- Some describe specific cases or examples

Include both, if they reflect the SAME reasoning

Do NOT exclude an argument just because it is more specific

---

### Same Reasoning vs Related Topic (CRITICAL)

Two arguments match ONLY if they share the SAME underlying reasoning.

Include:
- Arguments that support the same idea, even with different wording or examples

Exclude:
- Arguments that discuss a different reason, even if they are related to the same topic

Sharing the same topic is NOT enough — the reasoning must be the same

---

### Decision Rule

For each argument, ask:

“Does this argument support the SAME reason as the key point?”

- Yes → include it
- No → exclude it

Do not include arguments just because they are vaguely related.

---

### Final Coverage Check (Required)

Before submitting:

- Review the arguments you did NOT select
- Check if any express the SAME reasoning in a different or more specific way

Add them if they match
Keep them excluded if they introduce a different reason

---

### Final Goal

Your selection should be:

- **Complete** → no matching arguments are missed
- **Precise** → no non-matching arguments are included

Select ALL and ONLY the correct matches

---

## Summary
For each example:
- Select arguments that share the same main idea
- Explain your reasoning
- Rate how well the key point summarises them

---

## Qualification Test (Pre-Screen)

### Purpose

Before starting the main annotation task, you will complete a short qualification test.

This test ensures that you understand how to identify **coherent semantic groups of arguments** and
apply the task correctly.

Only participants who pass this test will proceed to the main annotation phase.

---

### What You Will Do

For each question, you will see:

- A **Topic** – the general discussion
- A **Key Point (KP)** – a short statement expressing a central idea
- A list of **Arguments**

---

### Your Task

Select the arguments that form a **coherent semantic group** that can be **jointly summarised by the key point**.

### How the Test is Evaluated

Your answers will be automatically evaluated based on how well they match the correct semantic grouping.

To pass:
- You must achieve a high level of accuracy across the questions

---

### Attempts

- You will have **up to 2 attempts**
- If you do not pass, you will not proceed to the main task

---

### Important Notice

This test is designed to ensure high-quality annotations.

Please take your time and carefully consider the meaning of each argument.

---

### Summary

To pass the test:
- Identify arguments that share the same **underlying idea**
- Avoid selecting loosely related arguments
- Focus on **coherence**, not keywords
\end{prompt}

Prompt for arguments reassignments:
\begin{prompt}
# Annotation Instructions: Key Point Matching Task

## Goal
In this task, you will evaluate how well a given **argument** matches a set of **key points**.

Your job is to select the **one key point** that best expresses the same idea as the argument,
or choose **“None”** if no key point matches.

---

## What You Will See

For each example, you will be given:

- **Topic** – the general subject
- **Argument** – a sentence expressing a specific opinion or reason
- **Key Points** – a list of short statements summarising common ideas

---

## Step 1: Select the Best Matching Key Point

Choose **one** option:

- The key point that expresses the **same main idea** as the argument
- **None**, if no key point matches the argument

---

## How to Decide

Ask yourself:

> *Does this key point express the same underlying idea or reasoning as the argument?*

### Select a key point if:
- It captures the **same claim or reasoning**
- The meaning is the same, even if wording differs
- The argument and key point could be paraphrases

### Do NOT select a key point if:
- It only shares **words** but not meaning
- It is **related to the topic** but expresses a different idea
- It focuses on a **different reason or aspect**

---

## When to Select “None”

Choose **“None”** if:

- No key point expresses the **same core idea** as the argument
- All options are **different in meaning**, even if related

**Do not force a match.**

---

## Important Guidelines

- Focus on **meaning**, not wording
- Select the **best match**, not just a related option
- Only one answer is allowed
- Do not rely on keyword overlap

---

## Examples

### Example 1 (Match)

**Argument:**
“Using bicycles instead of cars can reduce air pollution in cities.”

**Key Points:**
- A1: Cycling helps reduce pollution 
- A2: Cars are faster than bicycles
- A3: Public transport should be improved
- A4: None

**Correct answer:** A1
Both express the idea that cycling reduces pollution.

---

### Example 2 (Choose the best match)

**Argument:**
“Eating fresh fruit regularly can improve overall health.”

**Key Points:**
- A1: Fresh food is often more expensive
- A2: Fruit consumption improves health 
- A3: Grocery stores should lower prices
- A4: None

**Correct answer:** A2
A1 is related but does not express the same idea.

---

### Example 3 (No Match)

**Argument:**
“Installing solar panels requires a high initial investment.”

**Key Points:**
- A1: Solar energy is environmentally friendly
- A2: Renewable energy reduces emissions
- A3: Governments should promote green energy
- A4: None 

**Correct answer:** A4 (None)
All key points describe benefits, not cost.

---

## Summary

For each example:

- Read the **argument carefully**
- Compare it with all key points
- Select the **best matching key point**, or **None**
- Focus on **shared meaning and reasoning**, not wording

---
# Step 2: Provide a Brief Justification

After selecting your answer, briefly explain **why you chose that option**.

---

## What to Write

- Focus on the **main shared idea or reasoning** between the argument and the selected key point

---

## Good Justifications

A good justification clearly explains **why the meanings match (or do not match)**:

- “Both express that cycling helps reduce pollution.”
- “The key point matches the argument’s idea that fruit improves health.”
- “This option reflects the same concern about high costs.”

---

## Avoid

- Repeating the argument or key point without explanation
- Vague statements such as:
  - “They are similar”
  - “This seems correct”
- Focusing only on shared words instead of meaning

---

## If You Selected “None”

If you choose **“None”**, explain:

> *Why none of the key points match the argument*

### Example

- “The argument discusses financial cost, while all key points focus on environmental benefits.”

---

## Keep It Short and Clear

- Do **not** write long explanations
- Focus only on the **key reason for your decision**
- Clarity is more important than detail

---

## Summary

Your justification should answer:

> *Why does this key point match (or not match) the argument’s meaning?*

---

## Qualification Test (Pre-Screen)

## Purpose
Before starting the main task, you will complete a short pre-screen test.
This ensures that you understand how to correctly match **arguments** to **key points**.

---

## What You Will See

For each question, you will be given:

- **Topic** – the general subject
- **Argument** – a sentence expressing a specific idea
- **Key Points** – a list of possible matches

---

## Your Task

For each question:

- Select the **one key point** that best expresses the same idea as the argument
- OR select **“None”** if no key point matches

---

## How to Decide

Ask yourself:

> *Does this key point express the same meaning or reasoning as the argument?*

### Select a key point if:
- It captures the **same core idea**
- The meaning is the same, even if the wording is different

### Do NOT select a key point if:
- It only shares **similar words**
- It is **related to the topic** but expresses a different idea

---

## When to Select “None”

Choose **“None”** if:
- No key point expresses the **same idea** as the argument
- All options are only loosely related

Do **not** force a match.

---

## How You Will Be Evaluated

- Your answers will be compared to the correct selections
- You must achieve a sufficient score to pass

---

## Attempts

- You will have **up to 2 attempts**
- If you do not pass, you will not proceed to the main task

---

## Important Notes

- Only **one answer** is allowed per question
- Focus on **meaning**, not keywords
- Take your time—accuracy is more important than speed

---

## Summary

To pass the test:

-  Match the argument to the **correct key point**, or choose **None**
- Focus on **shared meaning and reasoning**
-  Avoid selecting options that are only loosely related
\end{prompt}

Instructions for KPA quality evaluation:
\begin{prompt}
# Human Evaluation Instructions

## Task Overview

You will be presented with:

1. A **topic**
2. A set of **sampled arguments**
3. Two candidate Key Point Analysis (KPA) structures:
   - **Structure A**
   - **Structure B**

Each structure contains:

- A set of key points
- Groups of arguments assigned to key points
- Any unmatched arguments

Your task is to evaluate the quality of both structures and determine which structure
provides the better overall representation of the argument collection.

---

## General Guidelines

- Focus on the **meaning** of arguments and key points rather than exact wording.
- Different arguments may express the same underlying idea using different language.
- Evaluate the structures based only on the information provided.
- Carefully review **both structures** before assigning scores.
- Use the scoring guidelines below to ensure consistency.
- If two structures appear equally good overall, select **Tie**.

---

# Evaluation Criteria

## 1. Semantic Grouping

### Definition

Evaluate how well the structure groups together arguments expressing the same
underlying idea while separating arguments expressing different ideas.

### Consider

- Do arguments within the same group express similar reasoning?
- Are different ideas separated into different groups?
- Are any arguments assigned to inappropriate groups?

### Scoring Guide

#### 5 — Excellent

- Almost all arguments within each group express the same underlying idea.
- Different ideas are clearly separated.
- There are very few or no incorrect assignments.

#### 4 — Good

- Most groups are coherent.
- A few arguments may be imperfectly assigned.
- Overall grouping quality is strong.

#### 3 — Fair

- Some groups are coherent, but others contain mixed ideas.
- Several assignments are questionable.
- The structure is understandable but imperfect.

#### 2 — Poor

- Multiple groups contain arguments expressing noticeably different ideas.
- Many assignments appear questionable.
- Group boundaries are unclear.

#### 1 — Very Poor

- Groups are largely incoherent.
- Arguments expressing unrelated ideas are frequently grouped together.
- The grouping does not reflect the underlying argument structure.

---

## 2. Key Point Quality

### Definition

Evaluate how well the key points summarize the arguments assigned to them.

### Consider

- Does the key point accurately represent the arguments in the group?
- Does it capture the main shared idea?
- Is it clear and representative?

### Scoring Guide

#### 5 — Excellent

- Key points accurately and concisely summarize their assigned arguments.
- They clearly capture the shared reasoning within the group.

#### 4 — Good

- Most key points summarize their groups well.
- Minor omissions or wording issues may exist.
- The central meaning is still captured correctly.

#### 3 — Fair

- Some key points only partially represent their arguments.
- Certain summaries are either too broad or too specific.
- The groups are still understandable overall.

#### 2 — Poor

- Several key points do not adequately summarize their arguments.
- Important aspects of the arguments are missing.
- Some summaries are misleading.

#### 1 — Very Poor

- Many key points fail to represent the assigned arguments.
- The relationship between key points and their groups is often unclear.
- Summaries are largely inaccurate.

---

## 3. Coverage

### Definition

Evaluate how well the set of key points captures the major ideas expressed in the argument collection.

### Consider

- Are the major themes represented?
- Are important ideas missing?
- Do the unmatched arguments suggest gaps in the structure?

### Scoring Guide

#### 5 — Excellent

- Nearly all major themes are represented.
- Few or no important ideas are missing.
- Coverage appears comprehensive.

#### 4 — Good

- Most major themes are represented.
- Only a small number of less important ideas are missing.

#### 3 — Fair

- Some major themes are represented, but others are missing or weakly represented.
- Coverage is adequate but incomplete.

#### 2 — Poor

- Multiple important themes are missing.
- Large parts of the argument collection are poorly represented.

#### 1 — Very Poor

- Many major themes are not represented.
- The structure fails to capture a substantial portion of the argument collection.

---

# Overall Preference

After reviewing both structures and assigning scores,
select the structure that provides the better overall representation of the argument collection.

### Consider

- Semantic Grouping
- Key Point Quality
- Coverage

### Options

- Structure A
- Tie
- Structure B

Select **Tie** only if the two structures are approximately equal in overall quality.

---

# Justification (Required)

Please explain the reason for your overall preference.

Your explanation should refer to one or more of the following:

- Quality of argument grouping
- Quality of key points
- Coverage of major themes
- Redundancy among key points
- Treatment of unmatched arguments

### Example

> I preferred Structure B because its argument groups were more coherent and
the key points better summarized the assigned arguments. Structure A left several important arguments unmatched,
while Structure B covered most of the major themes discussed in the argument collection.

Please provide a clear and meaningful explanation rather than a short statement such as "A is better" or "B is better."

---

## Final Reminder

Before submitting:

- Review both structures carefully.
- Make sure all scores have been assigned.
- Select an overall preference.
- Provide a justification for your choice.

---

## Qualification Test (Pre-Screen)

Before starting the annotation task, you must complete a short pre-screening test to
demonstrate your understanding of the task.

The test contains **4 questions**, each with **one correct answer**, covering:

- Identifying arguments with the same underlying idea
- Evaluating how well a key point summarizes arguments
- Assessing coverage of major themes
- Identifying arguments that do not fit a group

## Instructions

- Focus on the **meaning** of arguments rather than their exact wording.
- Read each question carefully before selecting an answer.
- Choose the answer that best reflects the relationships between the arguments and key points.

## Passing Requirement

You must answer **at least 3 out of 4 questions correctly** to proceed to the annotation task.

## Attempts

You may take the pre-screening test **up to two times**.

- **Pass:** You will gain access to the annotation task.
- **Fail twice:** You will not be eligible to continue.

Thank you for taking the time to complete this test carefully.
\end{prompt}

\end{document}